\documentclass{article}

\usepackage[preprint,nonatbib]{neurips_2021}

\usepackage[utf8]{inputenc} 
\usepackage[T1]{fontenc}    
\usepackage{hyperref}       
\usepackage{url}            
\usepackage{booktabs}       
\usepackage{amsfonts}       
\usepackage{nicefrac}       
\usepackage{microtype}      
\usepackage{xcolor}         

\usepackage{graphicx}
\usepackage{subfigure}
\usepackage{makecell}
\usepackage[justification=centering]{caption}
\usepackage{amsmath,amssymb,amsfonts}
\usepackage{float}
\usepackage{multirow}
\usepackage{epsfig}
\usepackage{algorithm}
\usepackage{algorithmic}
\usepackage{epsfig}
\usepackage{epstopdf}
\usepackage{multirow}
\usepackage{bm}
\usepackage{times}

\title{Incomplete Graph Representation and Learning via Partial Graph Neural Networks}

\author{%
  Bo Jiang \\
 School of Computer Science and Technology\\
  Anhui University\\
  \texttt{jiangbo@ahu.edu.cn} \\
  \And
  Ziyan Zhang \\
  School of Computer Science and Technology \\
  Anhui University \\
  \texttt{zhangziyanahu@163.com} \\
}

\begin{document}

\maketitle

\begin{abstract}
Graph Neural Networks (GNNs) are gaining increasing attention on graph data learning tasks in recent years.
However, in many applications, graph may be coming in an incomplete form where attributes of graph nodes are partially unknown/missing.
Existing GNNs are generally designed on complete graphs which can not deal with attribute-incomplete graph data directly.
%
To address this problem, we develop a novel \emph{partial aggregation} based GNNs, named Partial Graph Neural Networks (PaGNNs), for attribute-incomplete graph representation and learning.
 Our work is motivated by the observation that the neighborhood aggregation function in standard GNNs can be equivalently viewed as the neighborhood reconstruction formulation.
Based on it, we define two novel partial aggregation (reconstruction) functions on incomplete graph and derive PaGNNs for incomplete graph data learning.
 Extensive experiments on several datasets demonstrate the effectiveness and efficiency of  the proposed PaGNNs.
\end{abstract}

\section{Instruction}

Learning on graph data is an important problem in machine learning and data mining area.
Graph Neural Networks (GNNs)~\cite{GCN,monti2017geometric,graphsage,velickovic2017graph,DGI,CGNN} have been demonstrated great success in representation learning of graph data and have gained an increasing attention in recent years.
Given an input graph $G(\mathbf{A},\mathbf{H})$ where $\mathbf{A}\in \mathbb{R}^{n\times n}$ and  $\mathbf{H}\in \mathbb{R}^{n\times d}$ denote the adjacency matrix and node attributes respectively, the core aspect of GNNs is their \emph{message propagation scheme} in which each node updates its representation by aggregating the representations from its neighbors via a specific aggregation function~\cite{SMGCN,graphsage,GCN}.

\begin{figure*}[!htpb]
\centering
\includegraphics[width=0.75\textwidth]{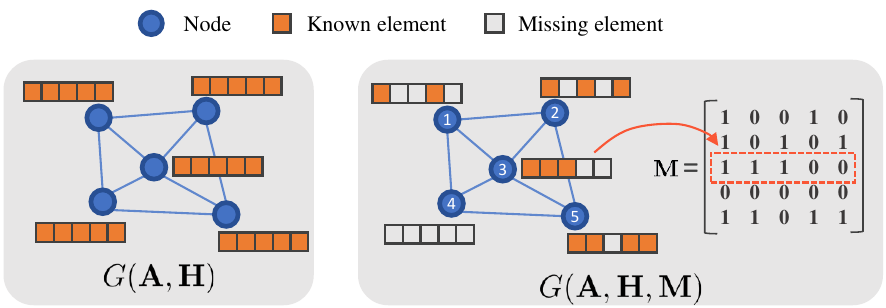}
  \caption{Illustration of complete graph $G(\mathbf{A},\mathbf{H})$ and attribute-incomplete graph $G(\mathbf{A},\mathbf{H},\mathbf{M})$.}
\label{fig:missing}
\end{figure*}

However, in many applications, graph data is usually coming in an incomplete form in which attributes of graph nodes are partially missing/unknown~\cite{GCNmf,SAT}. For example, some attributes are unknown or missing due to some difficulties or  human errors in data collection process~\cite{GCNmf}.
In this paper, we use $G(\mathbf{A},\mathbf{H},\mathbf{M})$ to represent this incomplete graph where  $\mathbf{M}\in \{0,1\}^{n\times d}$ denotes the indicative matrix in which   $\mathbf{M}_{ij}=0$ indicates that $\mathbf{H}_{ij}$ is missing/unknown and $\mathbf{M}_{ij}=1$ otherwise, as illustrated in Figure \ref{fig:missing}.
Existing GNNs are generally developed on attribute-complete graph data $G(\mathbf{A},\mathbf{H})$ which can not deal with incomplete graph data $G(\mathbf{A},\mathbf{H},\mathbf{M})$ directly.

\textbf{Related Works.}
To deal with attribute-incomplete graph $G(\mathbf{A},\mathbf{H},\mathbf{M})$, 
one popular way is to first employ some imputation/completion methods to estimate and fill in the missing values and then adopt traditional GNNs to conduct graph learning.
For example,
Yoon et al.~\cite{GAIN} propose Generative Adversarial Imputation Nets (GAIN) to complete missing data by adapting a GAN~\cite{GAN} framework.
Chen et al.~\cite{SAT} recently develop Structure-Attribute Transformer (SAT) for attribute-missing graph completion and then utilize GNNs~\cite{GCN,velickovic2017graph} for incomplete graph node classification.
Obviously, this kind of methods conduct missing value estimating and GNN learning independently, which usually leads to weak sub-optimal learning.
To overcome this limitation, Taguchi et al.~\cite{GCNmf} recently propose an end-to-end model GCN\scriptsize MF \normalsize which uses a Gaussian mixture model (GMM) to estimate the missing data and jointly learns the GMM and GNN's parameters together.
However, GCN\scriptsize MF \normalsize~\cite{GCNmf} increases the amount of training parameters and thus has high computational complexity.

\textbf{Contributions. }
Different from previous works, this paper presents a straightforward \textbf{partial message-propagation scheme} on incomplete graph  to develop novel Partial Graph Neural Networks (PaGNNs) for incomplete graph data learning.
Specifically, the proposed partial message propagation is motivated by the observation that the neighborhood aggregation operation in traditional GNNs can be equivalently viewed as the neighborhood reconstruction optimization.
Based on it, we derive a partial  reconstruction formulation and develop a novel partial  message-propagation scheme for GNN's layer-wise propagation. 
Overall, the main contributions of this paper are summarized as follows,
\begin{itemize}
\item We re-interpret neighborhood aggregation in GNNs from neighborhood reconstruction perspective and derive a novel partial aggregation function for GNN's layer-wise message propagation on incomplete graph.
\item We develop a simple and effective Partial Graph Neural Networks (PaGNNs)  to deal with incomplete graph data representation and learning. 
\end{itemize}

Experiments on several datasets demonstrate the effectiveness and efficiency  of the proposed PaGNNs.

\section{Revisiting GNNs}

%
%
It is known that the core aspect of GNNs is their \emph{message propagation scheme} in which each node updates its embedding by aggregating the embeddings from its neighbors via a specific aggregation function~\cite{SMGCN,graphsage,GCN,zhang2020feature,yang2019masked}.
Given an input graph $G(\mathbf{A}, \mathbf{H})$ where $\mathbf{H}=(\mathbf{h}_1, \mathbf{h}_2\cdots \mathbf{h}_n)\in \mathbb{R}^{n\times d}$, 
the message propagation 
for layer-wise propagation in GNNs can generally be formulated as,
\begin{flalign}\label{EQ:GNN}
\mathbf{h}_{i}' \gets \sigma\big(\phi_{agg}\left(\mathbf{{A}}_{ij},\mathbf{h}_{j},{j}\in \mathcal{N}_i\right) \mathbf{W}\big)
\end{flalign}
where $\phi_{agg}$ denotes the specific aggregation function and $\mathcal{N}_i$ denotes the $i$-th node's neighborhood.
$\mathbf{W}$ is network's weight parameter and $\sigma(\cdot)$ denotes an activation function. 
For example, one popular way to define the $\phi_{agg}$  is using weighted mean function~\cite{graphsage} as
\begin{flalign}\label{EQ:GCN_m}
\phi_{agg_m}\left(\mathbf{{A}}_{ij},\mathbf{h}_{j},{j}\in \mathcal{N}_i\right) =  \frac{1}{\mathbf{\tilde{d}}_i}\sum_{j\in \mathcal{N}_i}  \mathbf{\tilde{A}}_{ij}\mathbf{h}_{j}
\end{flalign}
where $\mathbf{\tilde{A}} = \mathbf{A}+\mathbf{I}$, ${\mathbf{\tilde{d}}}_{i}=\sum_{j_\in \mathcal{N}_i}{\mathbf{\tilde{A}}}_{ij}$ and $\mathbf{I}$ is an identity matrix.
In addition,
Kipf et al.~\cite{GCN} propose a commonly used Graph Convolutional Networks (GCN) which defines $\phi_{agg}$ as a symmetric normalized summation function as 
\begin{flalign}\label{EQ:GCN_n}
\phi_{agg_n}\left(\mathbf{{A}}_{ij},\mathbf{h}_{j},{j}\in \mathcal{N}_i\right) = \sum_{j\in \mathcal{N}_i}  \frac{1}{\sqrt{\mathbf{\tilde{d}}_i\mathbf{\tilde{d}}_j}} {\mathbf{\tilde{A}}}_{ij}\mathbf{h}_{j}
\end{flalign}
where $\mathbf{\tilde{A}} = \mathbf{A}+\mathbf{I}$ and ${\mathbf{\tilde{d}}}_{i}=\sum_{j_\in \mathcal{N}_i}{\mathbf{\tilde{A}}}_{ij}$.


\section{Partial Aggregation for Incomplete Graph Neural Networks}

The above standard aggregation functions ($\phi_{agg_m}, \phi_{agg_n}$) and message propagation scheme are defined on complete graph $G(\mathbf{A}, \mathbf{H})$, which limit standard GNNs~\cite{graphsage,GCN} to deal with incomplete graph data $G(\mathbf{A}, \mathbf{H}, \mathbf{M})$.
 To overcome this problem,
in this section, 
we define a novel \emph{partial} aggregation and message propagation scheme on incomplete graph $G(\mathbf{A}, \mathbf{H}, \mathbf{M})$ and propose Partial Graph Neural Networks (PaGNNs) for incomplete graph data learning.


Our partial aggregations are motivated based on re-formulations of the above neighborhood aggregations ($\phi_{agg_m}, \phi_{agg_n}$) from a neighborhood reconstruction perspective.
Specifically, one can easily prove that Eq.(\ref{EQ:GCN_m}) provides the optimal solution to
the following neighborhood reconstruction problem,
\begin{equation}\label{EQ:gnn_problem_m}
\begin{split}
\phi_{agg_m}\left(\mathbf{{A}}_{ij},\mathbf{h}_{j},{j}\in \mathcal{N}_i\right) =\arg\min_{\mathbf{z}} \sum_{j\in \mathcal{N}_i} \mathbf{\tilde{A}}_{ij}\| \mathbf{z} - \mathbf{h}_j\|^2
\end{split}
\end{equation}
where $\|\cdot\|$ denotes Frobenius norm~\cite{SMGCN} and $\mathbf{\tilde{A}} = \mathbf{A}+\mathbf{I}$. 

\textbf{Remark.} Eq.(\ref{EQ:gnn_problem_m}) provides a different view of the aggregation operation Eq.(\ref{EQ:GCN_m}) in GNN, i.e., each node aims to reconstruct its representation from its neighbors' representations.

Using this formulation,
we can derive a simple partial reconstruction (aggregation) function on incomplete graph $G(\mathbf{A},\mathbf{H},\mathbf{M})$ as
\begin{equation}\label{EQ:pagnn_problem_m}
\begin{split}
\phi_{pagg_m}\left(\mathbf{{A}}_{ij},\mathbf{h}_{j},\mathbf{m}_{j},{j}\in \mathcal{N}_i\right)= \arg\min_{\mathbf{z}}\sum_{j\in \mathcal{N}_i} \mathbf{\tilde{A}}_{ij}\|\mathbf{m}_j \odot \big(\mathbf{z} - \mathbf{h}_j\big)\|^2
\end{split}
\end{equation}
where $\mathbf{m}_j$ denotes the $j$-th row of indicative matrix $\mathbf{M}$ and $\odot$ denotes Hadamard product operation.
Moreover, Eq.(\ref{EQ:pagnn_problem_m}) also has a simple closed-form solution which is computed by setting the first derivation w.r.t $\mathbf{z}$ to $\mathbf{0}$ and obtain
\begin{equation}\label{EQ:pagnn_solution_m}
\begin{split}
\phi_{pagg_m}\left(\mathbf{{A}}_{ij},\mathbf{h}_{j},\mathbf{m}_{j},{j}\in \mathcal{N}_i\right) = \frac{\sum\nolimits_{j\in\mathcal{N}_i} \mathbf{\tilde{A}}_{ij}(\mathbf{m}_j\odot \mathbf{h}_j)}{\sum\nolimits_{j\in\mathcal{N}_i} \mathbf{\tilde{A}}_{ij} \mathbf{m}_j}
\end{split}
\end{equation}
Note that Eq.(\ref{EQ:pagnn_solution_m}) provides a partial neighborhood aggregation on (attribute) incomplete graph.
Using this partial aggregation function $\phi_{pagg_m}$, we can define a \emph{message propagation scheme} on incomplete graph as
 \begin{flalign}\label{EQ:pagnn_m}
\mathbf{h}_{i}' \gets \sigma\big(\phi_{pagg_m}\left(\mathbf{{A}}_{ij},\mathbf{h}_{j},\mathbf{m}_{j},{j}\in \mathcal{N}_i\right) \mathbf{W}\big)
\end{flalign}

Similarly, we can further prove that Eq.(\ref{EQ:GCN_n}) provides the optimal solution to problem as
\begin{equation}\label{EQ:gnn_problem_n}
\begin{split}
\phi_{agg_n}\left(\mathbf{{A}}_{ij},\mathbf{h}_{j},{j}\in \mathcal{N}_i\right) =
 \arg\min_{\mathbf{z}} \sum_{j\in \mathcal{N}_i} \mathbf{\hat{A}}_{ij}\Big{\|} \frac{\mathbf{z}}{\mathbf{\hat{d}}_i} - \mathbf{h}_j\Big{\|}^2
\end{split}
\end{equation}
where $\mathbf{\hat{d}}_i=\sum\nolimits_{j}\mathbf{\hat{A}}_{ij}$,
$\mathbf{\hat{A}}=\mathbf{\tilde{D}}^{-1/2}\mathbf{\tilde{A}}\mathbf{\tilde{D}}^{-1/2}$ and $\mathbf{\tilde{D}}$ is the diagonal matrix with $\mathbf{\tilde{D}}_{ii}=\sum\nolimits_{j}\mathbf{\tilde{A}}_{ij}$.
Using this formulation, we can also derive
a symmetric normalized type partial aggregation function on incomplete graph $G(\mathbf{A},\mathbf{H},\mathbf{M})$ as
\begin{equation}\label{EQ:pagnn_problem_n}
\begin{split}
\phi_{pagg_n}\left(\mathbf{{A}}_{ij},\mathbf{h}_{j},\mathbf{m}_{j},{j}\in \mathcal{N}_i\right)=\arg\min_{\mathbf{z}}\,\, \sum_{j\in \mathcal{N}_i} \mathbf{\hat{A}}_{ij}\Big{\|}\mathbf{m}_j \odot \big(\frac{\mathbf{z}}{\mathbf{\hat{d}}_i} - \mathbf{h}_j\big)\Big{\|}^2
\end{split}
\end{equation}
Eq.(\ref{EQ:pagnn_problem_n}) also has a simple closed-form solution as 
%
%
\begin{equation}\label{EQ:pagnn_solution_n}
\begin{split}
\phi_{pagg_n}\left(\mathbf{{A}}_{ij},\mathbf{h}_{j},\mathbf{m}_{j},{j}\in \mathcal{N}_i\right)= \frac{\mathbf{\hat{d}}_i\sum\nolimits_{j\in\mathcal{N}_i} \mathbf{\hat{A}}_{ij}(\mathbf{m}_j\odot \mathbf{h}_j)}{\sum\nolimits_{j\in\mathcal{N}_i} \mathbf{\hat{A}}_{ij} \mathbf{m}_j}
\end{split}
\end{equation}
Similar to Eq.(\ref{EQ:pagnn_m}), using this partial aggregation function $\phi_{pagg_n}$, we can define a normalized-type message propagation scheme on incomplete graph as
 \begin{flalign}\label{EQ:pagnn_n}
\mathbf{h}_{i}' \gets \sigma\big(\phi_{pagg_n}\left(\mathbf{{A}}_{ij},\mathbf{h}_{j},\mathbf{m}_{j},{j}\in \mathcal{N}_i\right) \mathbf{W}\big)
\end{flalign}
%


\textbf{Remark.}
Using the above message propagation operations ($\phi_{pagg_m}$ and $\phi_{pagg_n}$), we can develop two novel multi-layer GNNs for incomplete graph data learning. In this paper, we call them as Partial Graph Neural Networks (PaGNNs).
PaGNNs (PaGNN-M, PaGNN-N) have two main advantages.
 First, PaGNNs adopt standard GNNs-like message propagation scheme and thus can be implemented as simply and efficiently as standard GNNs.
In particular, both  Eq.(\ref{EQ:pagnn_m}) and Eq.(\ref{EQ:pagnn_n}) have GCN-like~\cite{GCN} matrix formulation as
\begin{align}
&\mathbf{H}' \gets \sigma\big[(\mathbf{\tilde{A}}(\mathbf{M}\odot \mathbf{H})\oslash\mathbf{\tilde{A}}\mathbf{M})\mathbf{W}\big] \label{EQ:rmgcn_matrix_solution} \\
&\mathbf{H}' \gets \sigma\big[(\mathbf{\hat{A}}\mathbf{E}\odot\mathbf{\hat{A}}(\mathbf{M}\odot \mathbf{H})\oslash\mathbf{\hat{A}}\mathbf{M})\mathbf{W}\big] \label{EQ:smgcn_matrix_solution}
\end{align}
where $\odot$ and $\oslash$ denote the Hadamard product and element division operation respectively and $\mathbf{E}\in \mathbb{R}^{n\times d}$ with all elements $\mathbf{E}_{ij}=1$.
$\mathbf{\tilde{A}}=\mathbf{A}+\mathbf{I}$,
$\mathbf{\hat{A}}=\mathbf{\tilde{D}}^{-1/2}\mathbf{\tilde{A}}\mathbf{\tilde{D}}^{-1/2}$ and $\mathbf{\tilde{D}}$ is the diagonal matrix with $\mathbf{\tilde{D}}_{ii}=\sum_{j}\mathbf{\tilde{A}}_{ij}$.
 Second, PaGNNs conduct on incomplete graph directly and do not need some additional parameters to estimate missing values. This is the key difference between our PaGNNs and previous related works~\cite{GAIN,GCNmf,SAT}.

\begin{figure*}[!htpb]
\centering
\includegraphics[width=0.8\textwidth]{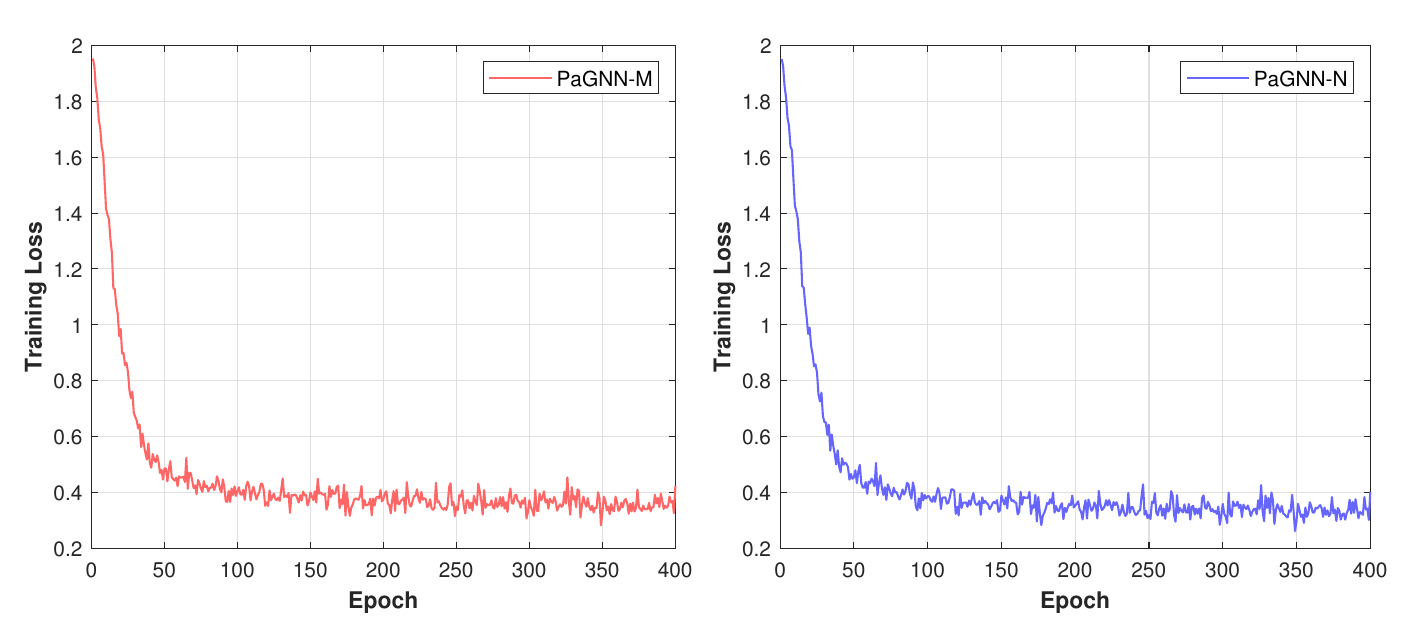}
  \caption{Demonstration of training loss values across different epochs on Cora dataset with 50\% unknown/missing attributes.}
\label{fig:loss}
\end{figure*}

The proposed PaGNNs are general which can be potentially used in many incomplete graph learning tasks, such as graph classification~\cite{deffpool}, link prediction~\cite{GAE} and clustering~\cite{cluster}.
In this paper, we mainly focus on semi-supervised node classification~\cite{GCN}.
Figure \ref{fig:loss} shows the cross-entropy classification loss of PaGNNs across different epochs by using well known Adam algorithm~\cite{Adam} on Cora\cite{sen2008collective} dataset with 50\% unknown/missing attributes (as introduced in Experiments in detail). It illustrates the convergence of the proposed PaGNNs (PaGNN-M (Eq.(\ref{EQ:rmgcn_matrix_solution})) and PaGNN-N (Eq.(\ref{EQ:smgcn_matrix_solution}))).


\section{Experiment}
We evaluate the effectiveness and efficiency of the proposed PaGNNs on four datasets and compare them against several related  approaches.

\begin{table*}[!htp]
\centering
\renewcommand\arraystretch{1.5}
\caption{Statistics of datasets.}
\label{tab:dataset}
\centering
\small{
\begin{tabular}{c|c|c|c|c}
  \hline
  \hline
 &\#Nodes&\#Edges&\#Attributes&\#Classes\\
 \hline
 Cora~\cite{sen2008collective}&2708&5429&1433&7\\
 Citeseer~\cite{sen2008collective}&3327&4732&3703&6\\
 AmaPhoto~\cite{mcauley2015image} &7650&143663&745&8\\
 AmaComp~\cite{mcauley2015image} &13752&287209&767&10\\
 \hline
 \hline
\end{tabular}}
\end{table*}
\subsection{Experimental Setup}

We test the proposed PaGNNs on several commonly used datasets including  Cora~\cite{sen2008collective}, Citeseer~\cite{sen2008collective}, AmaPhoto~\cite{mcauley2015image} and AmaComp~\cite{mcauley2015image}.
The introductions of these datasets are summarized in Table \ref{tab:dataset}.
\begin{figure*}[!htpb]
\centering
\includegraphics[width=0.6\textwidth]{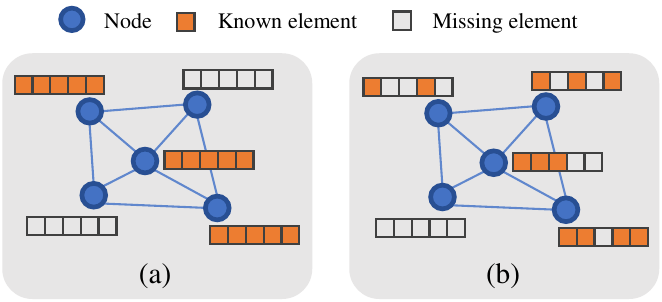}
  \caption{The illustration of two types of incomplete graph.}
\label{fig:twocase}
\end{figure*}
For  Cora and Citeseer datasets, similar to the standard data splits~\cite{GCN,GCNmf},
 we randomly choose 20 nodes per each class for training, 500 nodes for validation, 1000 nodes for testing.
For  AmaPhoto and AmaComp datasets, we randomly select 40 nodes per each class for training, 500 nodes for validation and the remaining nodes for testing, as used in works~\cite{GCNmf}.
For each dataset, we generate two types of attribute incomplete graph data as follows,
\begin{itemize}
\item Type 1: We randomly select $p_m$ percentage of graph nodes and set all attributes of these nodes to unknown/missing values, as illustrated in Figure \ref{fig:twocase} (a).
\item Type 2: We randomly select and set $p_m$ percentage of elements in attribute  matrix $\mathbf{H}$ to unknown/missing elements, as illustrated in in Figure \ref{fig:twocase} (b).
\end{itemize}
Similar to work~\cite{GCNmf}, for each missing rate $p_m$, we generate five different incomplete graphs and report the average learning accuracy.
For all compared methods, we report the average performance under 20 different network initializations, as suggested in work~\cite{GCNmf}.
Similar to the setting of standard GCN~\cite{GCN}, we use a two-layer architecture in our PaGNNs.
For node classification task, we optimize all network parameters by minimizing the cross-entropy loss function~\cite{GCN}.
The learning and dropout rates  of PaGNNs are set to $0.005$ and $0.5$, respectively.
The number of hidden units is set to 16 on Cora and Citeseer datasets and 64 on AmaPhoto and AmaComp datasets.
All network parameters are optimized by using Adam algorithm\cite{Adam} in which the training is stopped if the validation loss does not decrease for 100 epochs, as utilized in many previous works~\cite{GCNmf,GCN}.

\subsection{Comparison Methods}


We first compare our PaGNNs with some commonly used imputation based methods which include Mean Filling (MF)~\cite{liu2019multiple}, VAE~\cite{vae} and GAIN~\cite{GAIN}. These methods first estimate and fill in the missing values for the incomplete graph and then use GCN~\cite{GCN}  on the `filled' complete graph.
We also compare PaGNNs against some recent competitive methods  including Structure-Attribute Transformer (SAT)\cite{SAT} and GCN\scriptsize MF \normalsize\cite{GCNmf}.
The detail of these comparison methods are introduced below.
\begin{itemize}
  \item MF~\cite{liu2019multiple} fills in all the missing values in attributes $\mathbf{H}$ by using the mean value of the known elements across different nodes.
   \item VAE~\cite{vae}  utilizes a VAE model to recover the missing values in attributes $\mathbf{H}$.
   \item GAIN~\cite{GAIN}  uses a generative adversarial network (GAN)~\cite{GAN} technique to generate the missing values in attributes $\mathbf{H}$.
   \item SAT (GCN)~\cite{SAT}  first estimates the missing attributes via a GNN-based distribution matching model and then uses standard GNNs~\cite{GCN,velickovic2017graph} for incomplete graph data learning. 
   \item GCN\scriptsize MF \normalsize\cite{GCNmf}  employs a learnable Gaussian mixture model (GMM) to fill in the missing data and learns all parameters in an end-to-end manner.
\end{itemize}
%
\begin{table*}[!htp]
\centering
\renewcommand\arraystretch{1.5}
\caption{Comparison results on Type 1 problem under different missing rates.
$^{\ast}$ means that the results are obtained from previous work~\cite{GCNmf}.}
\label{tab:result1}
\centering
\small{
\begin{tabular}{c|c|c|c|c|c|c|c|c|c|c}
  \hline
  \hline
\multirow{2}{*}{Method}&\multicolumn{5}{c|}{Cora}&\multicolumn{5}{c}{Citeseer}\\
\cline{2-11}
  &0.1 & 0.3 &0.5 & 0.7&0.9&0.1 & 0.3 &0.5 & 0.7&0.9 \\
\hline
MF\cite{liu2019multiple}               &81.03&79.53&76.67&67.80&35.13&70.27&68.87&63.23&55.67&27.30\\
VAE$^{\ast}$\cite{vae}                 &80.63&78.57&74.69&60.71&17.27&69.63&66.34&60.46&40.71&17.20\\
GAIN$^{\ast}$\cite{GAIN}               &80.53&78.36&74.25&61.33&18.43&69.47&65.88&59.96&41.21&17.89\\
SAT\cite{SAT}                    &81.14&80.11&76.64&71.48&51.67&70.33&69.08&65.12&\textbf{58.52}&34.32\\
GCN\scriptsize MF \normalsize$^{\ast}$~\cite{GCNmf} &81.65&80.67&77.43&\textbf{72.69}&55.64&70.44&66.57&63.44&56.88&39.86\\
\cline{1-11}
PaGNN-M                                &81.40&80.52&\textbf{77.54}&68.94&48.78&70.84&68.84&65.02&56.86&41.14\\
PaGNN-N                                &\textbf{81.78}&\textbf{80.82}&77.46&72.32&\textbf{56.60}&\textbf{71.44}&\textbf{69.84}&\textbf{66.56}&58.34&\textbf{42.42}\\
 \hline
 \hline
 \multirow{2}{*}{Method}&\multicolumn{5}{c|}{AmaPhoto}&\multicolumn{5}{c}{AmaComp}\\
\cline{2-11}
  &0.1 & 0.3 &0.5 & 0.7&0.9&0.1 & 0.3 &0.5 & 0.7&0.9 \\
\hline
MF\cite{liu2019multiple}               &91.77&91.07&90.03&88.87&81.53&84.17&83.50&82.27&79.00&51.40\\
VAE$^{\ast}$\cite{vae}                 &92.11&91.50&90.46&87.47&67.85&82.76&81.72&79.23&73.76&41.37\\
GAIN$^{\ast}$\cite{GAIN}               &92.04&91.49&90.63&88.60&76.48&82.76&82.11&80.76&74.38&54.20\\
SAT\cite{SAT}                    &92.10&91.03&90.45&89.99&88.98&83.48&82.66&82.40&80.62&80.17\\
GCN\scriptsize MF \normalsize$^{\ast}$~\cite{GCNmf} &92.45&92.08&91.52&90.39&86.09&86.37&85.80&85.24&84.06&73.42\\
\cline{1-11}
PaGNN-M                                &92.48&92.24&91.70&91.12&89.08&\textbf{86.76}&86.26&\textbf{85.86}&85.04&82.70\\
PaGNN-N                                &\textbf{92.74}&\textbf{92.42}&\textbf{91.96}&\textbf{91.52}&\textbf{89.12}&86.40&\textbf{86.28}&85.84&\textbf{85.36}&\textbf{83.16}\\
 \hline
 \hline
\end{tabular}}
\end{table*}

\begin{table*}[!htp]
\centering
\renewcommand\arraystretch{1.5}
\caption{Comparison results on Type 2 problem under different missing rates.
$^{\ast}$ means that the results are obtained from previous work~\cite{GCNmf}. }
\label{tab:result2}
\centering
\small{
\begin{tabular}{c|c|c|c|c|c|c|c|c|c|c}
  \hline
  \hline
\multirow{2}{*}{Method}&\multicolumn{5}{c|}{Cora}&\multicolumn{5}{c}{Citeseer}\\
\cline{2-11}
  &0.1 & 0.3 &0.5 & 0.7&0.9&0.1 & 0.3 &0.5 & 0.7&0.9 \\
\hline
MF\cite{liu2019multiple}               &80.40&78.07&74.37&66.70&37.40&70.87&69.90&66.27&59.83&39.17\\
VAE$^{\ast}$\cite{vae}                 &80.91&79.18&76.84&50.79&13.27&69.80&68.54&50.91&18.45&11.00\\
GAIN$^{\ast}$\cite{GAIN}               &80.43&78.35&75.31&70.34&58.87&69.64&67.56&63.86&55.77&42.73\\
GCN\scriptsize MF \normalsize$^{\ast}$~\cite{GCNmf} &81.70&80.41&77.91&74.38&63.49&70.93&69.84&67.03&60.70&47.78\\
\cline{1-11}
PaGNN-M                                &82.02&80.56&77.96&74.90&65.16&70.84&69.88&67.20&61.34&48.50\\
PaGNN-N                                &\textbf{82.22}&\textbf{80.64}&\textbf{78.88}&\textbf{75.40}&\textbf{66.88}&\textbf{71.38}&\textbf{69.98}&\textbf{67.46}&\textbf{63.88}&\textbf{52.46}\\
 \hline
 \hline
 \multirow{2}{*}{Method}&\multicolumn{5}{c|}{AmaPhoto}&\multicolumn{5}{c}{AmaComp}\\
\cline{2-11}
  &0.1 & 0.3 &0.5 & 0.7&0.9&0.1 & 0.3 &0.5 & 0.7&0.9 \\
\hline
MF\cite{liu2019multiple}               &91.83&91.50&90.77&89.47&79.83&83.97&83.60&82.67&81.23&62.40\\
VAE$^{\ast}$\cite{vae}                 &92.20&91.90&91.15&89.28&81.43&82.65&81.72&80.47&78.55&67.26\\
GAIN$^{\ast}$\cite{GAIN}               &92.23&91.90&91.49&90.72&86.96&82.94&82.44&81.56&79.96&76.15\\
GCN\scriptsize MF \normalsize$^{\ast}$~\cite{GCNmf} &92.54&92.20&92.09&91.25&88.96&86.32&85.98&85.46&84.03&77.52\\
\cline{1-11}
PaGNN-M                                &92.64&92.48&92.14&91.74&90.48&\textbf{86.80}&\textbf{86.54}&\textbf{86.14}&85.32&83.88\\
PaGNN-N                                &\textbf{92.78}&\textbf{92.80}&\textbf{92.40}&\textbf{91.98}&\textbf{90.82}&86.38&86.10&85.78&\textbf{85.46}&\textbf{84.52}\\
 \hline
 \hline
\end{tabular}}
\end{table*}
\begin{figure*}[!htpb]
\centering
\includegraphics[width=1\textwidth]{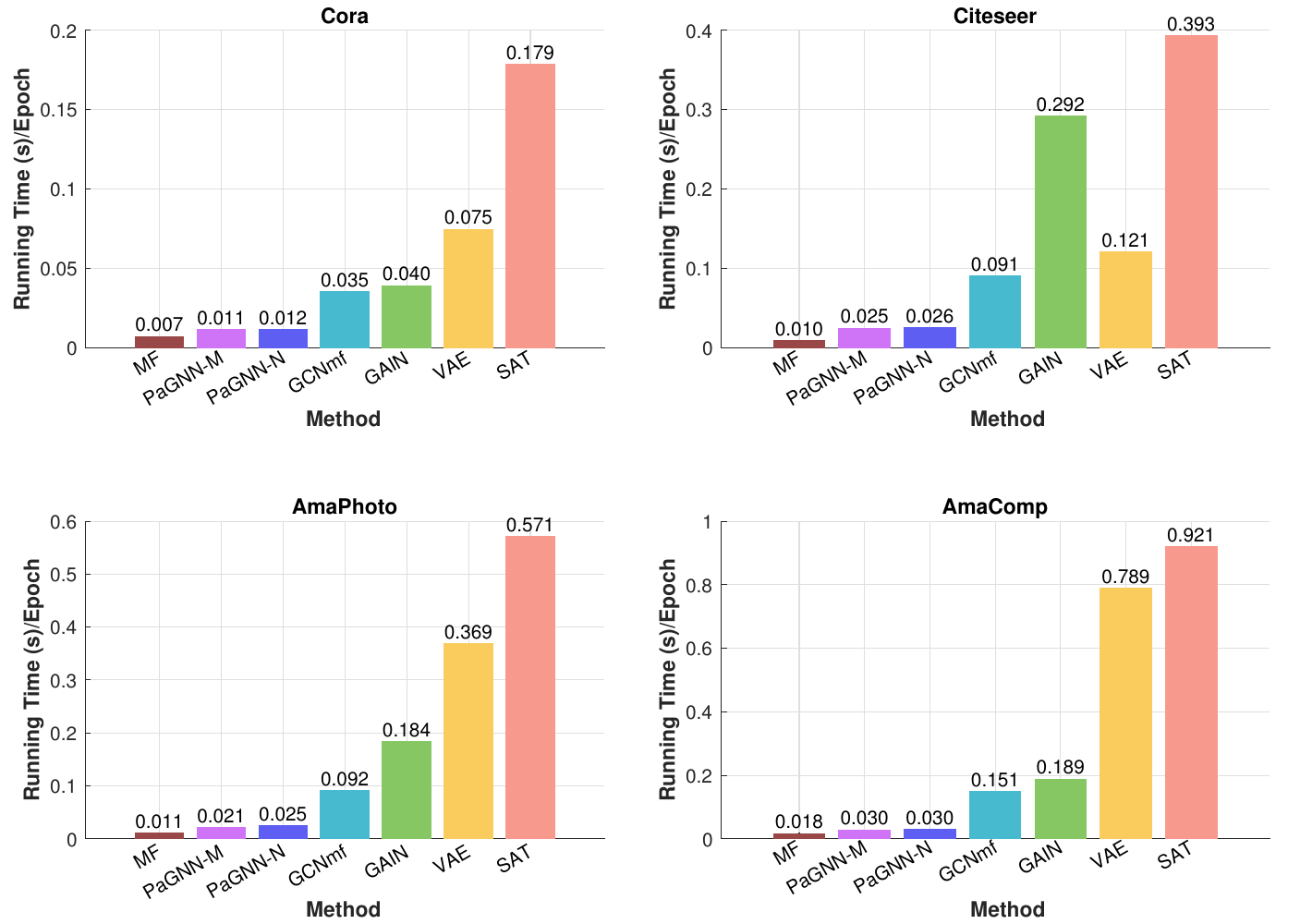}
  \caption{The empirical running time of each epoch for different comparison methods.}
\label{fig:time}
\end{figure*}

\subsection{Comparison Results}

\textbf{Effectiveness Evaluation.}  Table \ref{tab:result1} and Table \ref{tab:result2} summarize the comparison results on problem Type 1 and Type 2, respectively.
Since SAT~\cite{SAT} can only deal with problem Type 1, we have not compared it in Table \ref{tab:result2}.
Overall, we can note that  PaGNNs generally perform better than the other comparison methods, especially on the incomplete graph data with large number of missing values.
Specifically,  we can observe that,
(1) PaGNNs perform better than some traditional imputation based methods including MF~\cite{liu2019multiple}, VAE~\cite{vae} and GAIN~\cite{GAIN} on both types of incomplete graph data.
(2) PaGNNs generally obtain better performance than recent SAT~\cite{SAT} method for Type 1 problem. 
(3) PaGNNs outperform recent GCN\scriptsize MF \normalsize\cite{GCNmf} method which also provides an end-to-end model for incomplete graph data learning.
Comparing with GCN\scriptsize MF \normalsize\cite{GCNmf}, PaGNNs deal with incomplete graph directly without needing some additional parameters to estimate the unknown information and thus perform more optimally.
(4) PaGNN-N generally performs slightly better than PaGNN-M, indicating the more effectiveness of the proposed normalized aggregation function $\phi_{pagg_n}$ in Eq.(\ref{EQ:pagnn_solution_n})

\textbf{Efficiency Evaluation.}
Figure \ref{fig:time} shows the running time of each training epoch of PaGNNs on different datasets which are implemented on the computer with NVIDIA GeForce GTX 1080 Ti GPU.
For fair comparison, the average running time of each compared two-stage method (MF~\cite{liu2019multiple}, GAIN~\cite{GAIN} , VAE~\cite{vae}, SAT\cite{SAT}) is calculated as the whole time cost (missing value estimation + GCN training) divided by total epochs on GCN training.
Here, we can observe that
(1) PaGNNs are slightly slower than MF~\cite{liu2019multiple} method which uses a simple mean strategy for data imputation and standard GCN~\cite{GCN} for graph learning.
As shown in Eqs.(\ref{EQ:rmgcn_matrix_solution},\ref{EQ:smgcn_matrix_solution}), PaGNNs have GCN-like message propagation operation and thus can have similar time complexity (slightly higher) with standard GCN~\cite{GCN}.
(2) PaGNNs perform obviously faster than other comparing methods including GAIN~\cite{GAIN} , VAE~\cite{vae}, SAT\cite{SAT} and GCN\scriptsize MF \normalsize\cite{GCNmf}.
This clearly demonstrates the efficiency of the proposed PaGNNs.

%
%

\section{Conclusion}

In this paper, we develop a novel PaGNNs for attribute-incomplete graph data learning and representation.
PaGNNs adopt GCN-like message propagation scheme and thus can be implemented as simply and efficiently as standard GNNs.
PaGNNs conduct on incomplete graph directly and do not need some additional parameters to estimate missing values.
Extensive experiments show the superiority of our PaGNNs against other comparison methods.


\begin{thebibliography}{10}

\bibitem{GCN}
T.~N. Kipf and M.~Welling, ``Semi-supervised classification with graph
  convolutional networks,'' in {\em International Conference on Learning
  Representations (ICLR)}, 2017.

\bibitem{monti2017geometric}
F.~Monti, D.~Boscaini, J.~Masci, E.~Rodola, J.~Svoboda, and M.~M. Bronstein,
  ``Geometric deep learning on graphs and manifolds using mixture model cnns,''
  in {\em IEEE conference on computer vision and pattern recognition (CVPR)},
  pp.~5115--5124, 2017.

\bibitem{graphsage}
W.~L. Hamilton, Z.~Ying, and J.~Leskovec, ``Inductive representation learning
  on large graphs,'' in {\em Advances in Neural Information Processing Systems
  (NIPS)}, pp.~1024--1034, 2017.

\bibitem{velickovic2017graph}
P.~Veli{\v{c}}kovi{\'c}, G.~Cucurull, A.~Casanova, A.~Romero, P.~Li{\`o}, and
  Y.~Bengio, ``Graph attention networks,'' in {\em International Conference on
  Learning Representations (ICLR)}, 2018.

\bibitem{DGI}
P.~Veli{\v{c}}kovi{\'c}, W.~Fedus, W.~L. Hamilton, P.~Li{\`o}, Y.~Bengio, and
  R.~D. Hjelm, ``Deep graph infomax,'' in {\em International Conference on
  Learning Representations (ICLR)}, 2019.

\bibitem{CGNN}
Z.~Xinyi and L.~Chen, ``Capsule graph neural network,'' in {\em International
  Conference on Learning Representations (ICLR)}, 2018.

\bibitem{SMGCN}
S.~Geisler, D.~Z{\"u}gner, and S.~G{\"u}nnemann, ``Reliable graph neural
  networks via robust aggregation,'' in {\em Advances in Neural Information
  Processing Systems (NIPS)}, pp.~13272--13284, 2020.

\bibitem{GCNmf}
H.~Taguchi, X.~Liu, and T.~Murata, ``Graph convolutional networks for graphs
  containing missing features,'' {\em Future Generation Computer Systems},
  vol.~117, pp.~155--168, 2021.

\bibitem{SAT}
X.~Chen, S.~Chen, J.~Yao, H.~Zheng, Y.~Zhang, and I.~W. Tsang, ``Learning on
  attribute-missing graphs,'' {\em IEEE Transactions on Pattern Analysis and
  Machine Intelligence}, 2020.

\bibitem{GAIN}
J.~Yoon, J.~Jordon, and M.~Schaar, ``Gain: Missing data imputation using
  generative adversarial nets,'' in {\em International Conference on Machine
  Learning (ICML)}, pp.~5689--5698, PMLR, 2018.

\bibitem{GAN}
I.~J. Goodfellow, J.~Pouget-Abadie, M.~Mirza, B.~Xu, D.~Warde-Farley, S.~Ozair,
  A.~Courville, and Y.~Bengio, ``Generative adversarial networks,'' in {\em
  Advances in Neural Information Processing Systems (NIPS)}, pp.~2672--2680,
  2014.

\bibitem{zhang2020feature}
L.~Zhang and H.~Lu, ``A feature-importance-aware and robust aggregator for
  gcn,'' in {\em ACM International Conference on Information \& Knowledge
  Management (CIKM)}, pp.~1813--1822, 2020.

\bibitem{yang2019masked}
L.~Yang, F.~Wu, Y.~Wang, J.~Gu, and Y.~Guo, ``Masked graph convolutional
  network.,'' in {\em International Joint Conference on Artificial Intelligence
  (IJCAI)}, pp.~4070--4077, 2019.

\bibitem{deffpool}
R.~Ying, J.~You, C.~Morris, X.~Ren, W.~L. Hamilton, and J.~Leskovec,
  ``Hierarchical graph representation learning with differentiable pooling,''
  in {\em Advances in Neural Information Processing Systems (NIPS)},
  pp.~4805--4815, 2018.

\bibitem{GAE}
T.~N. Kipf and M.~Welling, ``Variational graph auto-encoders,'' in {\em
  Advances in Neural Information Processing Systems (NIPS) Bayesian Deep
  Learning Workshop}, 2016.

\bibitem{cluster}
J.~Xie, G.~Ross, and A.~Farhadi, ``Unsupervised deep embedding for clustering
  analysis,'' in {\em International Conference on Machine Learning (ICML)},
  pp.~478--487, 2016.

\bibitem{Adam}
D.~P. Kingma and J.~Ba, ``Adam: A method for stochastic optimization,'' in {\em
  International Conference on Learning Representations (ICLR)}, 2015.

\bibitem{sen2008collective}
P.~Sen, G.~Namata, M.~Bilgic, L.~Getoor, B.~Galligher, and T.~Eliassi-Rad,
  ``Collective classification in network data,'' {\em AI magazine}, vol.~29,
  no.~3, p.~93, 2008.

\bibitem{mcauley2015image}
J.~McAuley, C.~Targett, Q.~Shi, and A.~Van Den~Hengel, ``Image-based
  recommendations on styles and substitutes,'' in {\em Special Interest Group
  on Information Retrieval (SIGIR)}, pp.~43--52, 2015.

\bibitem{liu2019multiple}
X.~Liu, X.~Zhu, M.~Li, L.~Wang, E.~Zhu, T.~Liu, M.~Kloft, D.~Shen, J.~Yin, and
  W.~Gao, ``Multiple kernel k-means with incomplete kernels,'' {\em IEEE
  transactions on pattern analysis and machine intelligence}, 2019.

\bibitem{vae}
D.~Kingma and M.~Welling, ``Auto-encoding variational bayes,'' in {\em
  International Conference on Learning Representations (ICLR)}, 2014.

\end{thebibliography}
\end{document}